\pdfoutput=1
\documentclass[11pt]{article}
\usepackage[final]{acl}
\usepackage{times}
\usepackage{latexsym}
\usepackage[T1]{fontenc}
\usepackage[utf8]{inputenc}
\usepackage{microtype}
\usepackage{inconsolata}
\usepackage{graphicx}
\usepackage{booktabs}
\usepackage{amsmath}
\usepackage{amssymb}

\title{Fidelity Collapses with Table Size, Not Turn Count: Delivery
  Architecture in a Deployed Tool-Using Agent}

\author{Tung Hoang \\
  AwardWise, San Jose, USA \\
  \texttt{hoangthanhtungvn@gmail.com}}

\begin{document}
\maketitle

\begin{abstract}
When a multi-turn agent delivers a tool's structured records into the
model's context and re-emits them each turn, every turn can corrupt data
the tool computed correctly. We measure this in a production award-travel
agent: two architectures over identical frozen payloads, lineage-matched
prompts and sampling, differing only in delivery architecture. \textbf{SYNTHESIS} feeds
full record sets into the planner's context and answers follow-ups in-head;
\textbf{BYPASS}, the deployed system, keeps records in session state,
translates follow-ups into arguments for a deterministic filter, and
delivers rows over a typed side channel. Across 24 dialogues (339 scored
turns per condition), BYPASS scores 92.3\% of turns fully correct,
SYNTHESIS 45.7\% (paired gap \textbf{+46.6pp} $[+35.7, +57.1]$). Varying delivered-table size by capture date, SYNTHESIS collapses
from near-perfect at ${\le}10$ rows to ${\le}8\%$ beyond ${\sim}74$
rows. On a content-controlled prefix ladder the collapse threshold orders by model capability, from under 10 rows (GPT-4o-mini) to ${\gtrsim}198$ (Gemini 2.5 Pro), while BYPASS stays size-flat for all
four models. The collapse is expensive: beyond 100 rows, where almost no SYNTHESIS answer is correct, each correct one costs roughly 200$\times$ BYPASS's flat \$0.012. Practitioners converged on typed side-channel
delivery for rendering, not correctness; we show it was, unknowingly, also a correctness decision.
\end{abstract}

\section{Introduction}
\label{sec:intro}

Structured hallucination, in which an agent retrieves correct structured data
through a tool call and then corrupts field values while formatting them, is invisible to ordinary hallucination detection \citep{ji2023survey}:
validators see valid JSON, schema-parse metrics log success
\citep{hoang2026structured}. Prior work established the phenomenon
\textbf{single-turn}; a deployed award-travel session runs five to twelve
turns, each another synthesis pass over the same record set. Whether the
corruption compounds, and what it scales with, is this paper's question.

\begin{figure}[t]
  \centering
  \includegraphics[width=\columnwidth]{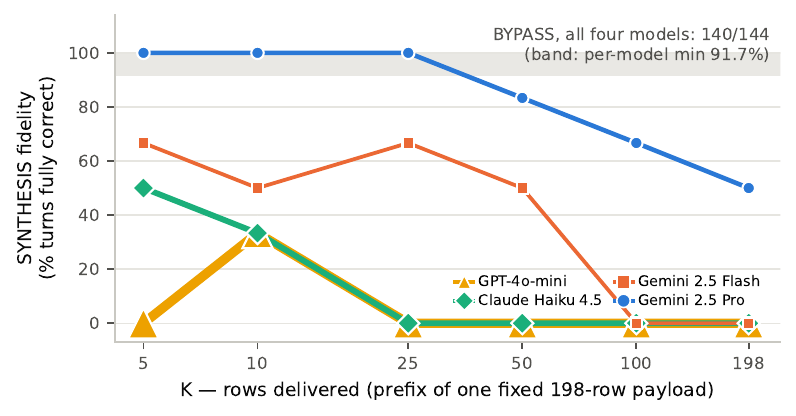}
  \caption{Four models, one content-controlled ladder: SYNTHESIS fidelity vs K-row prefixes of one frozen payload. \textbf{Capability buys threshold, architecture buys immunity}: collapse orders by capability (under 10 $\rightarrow$ ${\gtrsim}198$ rows); the BYPASS band is size-flat for all four (140/144 runs). $n{=}6$ runs per (model, K) cell.}
  \label{fig:family}
\end{figure}

The stake is architectural: agent frameworks default to returning a
tool's result into the model's context for the model to re-emit: every framework we surveyed does so, or requires wiring an opt-in bypass
\citep{langgraph2026docs,vercel2026docs,crewai2026docs,msagent2026docs,openai2026docs}.
The alternative (records as structured state, follow-ups answered by a
deterministic tool, rows streamed on a typed side channel) exists
\citep{copilotkit2026genui}, adopted for rendering, with correctness only
an unmeasured aside \citep{anthropic2025mcp,copilotkit2026issue}; the
corruption itself is documented both single-turn \citep{yang2026tables}
and, over long agentic workflows, at roughly a quarter of document content
\citep{laban2026corrupt}. To our knowledge, no prior work joins the two: a controlled A/B on
the same real record sets, with row- and field-level fidelity on
\textbf{both} arms. We measure the corruption in a production multi-agent award-and-cash flight system built on Google's Agent Development Kit (ADK) as a fork of the
\texttt{adk-samples} travel-concierge \citep{adksamples2025}: the
fork-point architecture (full row lists into the planner's context, follow-ups in-head) is a real system contract, not a strawman, and the
deployed system is the same tree with only delivery replaced; both run over the same frozen vendor payloads.

\paragraph{Contributions.}
\begin{itemize}
\item \textbf{A controlled size manipulation, measured on both arms of a production system, testing an architecture the field adopted on faith.}
  Typed side-channel delivery shipped for UX, never correctness
  (LangGraph in 2024, AG-UI in 2025, Vercel Data Parts in 2025, Google A2UI in 2026).
  Twenty dialogues over four routes vary delivered size
  (4$\rightarrow$194 rows); a six-route ladder holds \emph{content}
  constant via K-row prefixes (Figure~\ref{fig:topk}); a $3{\times}3$ Latin square separates size
  from position ($p=3{\times}10^{-10}$ vs $0.36$). SYNTHESIS collapses
  with size on every design; BYPASS is flat at \textbf{99.2\%}; a
  byte-identical control table holds its level even where treatment
  scored zero. Strict per-turn field fidelity gives \textbf{BYPASS
  92.3\% vs SYNTHESIS 45.7\%, paired gap +46.6pp}, replicating across three model families at +46.6 to +55.8pp: capability buys threshold, architecture buys immunity (Figure~\ref{fig:family}).
\item \textbf{A cost and latency account, not only an accuracy one.}
  BYPASS's per-turn cost and latency are flat in table size; SYNTHESIS's
  scale with it, reaching \textbf{\$2.53 per correct answer} beyond 100 rows against BYPASS's flat \$0.012 (Figure~\ref{fig:cpc}).
  Swapping the synthesis model from the production Flash to Gemini 2.5 Pro over the same 24 dialogues lifts synthesis fidelity from Flash's 46\% to 72\%, at 5$\times$ Flash's per-turn cost. Renting
  fidelity from a bigger model is the expensive substitute for not
  re-emitting the data.
\item \textbf{A published negative result, with diagnosis, and evidence
  that failures are not independent.} Our pre-registered test for
  super-linear error compounding against the accumulation-only null $1-(1-p)^n$ finds none, because the null saturates when per-turn failure is
  large. Its replacement, lag-1 conditional hazard (does this turn's
  failure predict the next turn's? \S\ref{sec:cluster}), agrees with
  value-level tracing: synthesis failures propagate, each corrupted
  emission becoming the next turn's input. BYPASS recomputes every turn from stored state; propagation is precluded by the delivery path, and the 0/25 reflects that no counterexample arose.
\item \textbf{A measurement methodology with computed ground truth, and
  its integrity report.} Expert-demonstrated dialogues replay against
  frozen vendor payloads with \emph{computed} ground truth
  (production filter evaluator; 408 frozen captures; detail in \S\ref{sec:method}). Practising it
  surfaced eight instrument defects (three in our own ground truth, \S\ref{sec:gtdefects}), plus two ground-truth semantics corrections and three production bugs, all fixed; the headline holds under both
  instrument versions.
\end{itemize}

\section{Related Work}
\label{sec:related}

\paragraph{Structured output: validity is not correctness.} Constrained
decoding guarantees shape, not content: validity hits 100\% while
accuracy falls and wrong-but-valid outputs dominate
\citep{ray2026constraint}. \citet{hoang2026structured} measured the
resulting structured hallucination in a production travel agent
single-turn, introducing the field-fidelity metric and zero-out policy we
adopt. This paper compares two delivery architectures on a new dataset in
a multi-turn design (results are not compared to that study's).

\paragraph{Multi-turn agent evaluation.} $\tau$-bench established
multi-turn tool-agent evaluation \citep{yao2024taubench}; a 39\% average
single- to multi-turn drop with poor recovery is documented
\citep{laban2025lost}; self-conditioning, in which past errors raise
later error rates, is the closest mechanism match
\citep{sinha2025illusion}. Closest to us, ``Beyond Task Completion''
reports 27--78\% of $\tau$-bench successes as \emph{corrupt}
\citep{cao2026corrupt}: procedural integrity on $\tau$-bench's mock domains (airline and retail),
where ours is field- and row-level data fidelity on real records, indexed
by delivered record-set size, with the corruption self-inflicted.

\paragraph{Architectures that keep data out of the model.} CodeAct routes
actions through executable code \citep{wang2024codeact}; practitioner
channels exist (LangGraph's custom stream mode, the Vercel AI SDK's data
parts, the AG-UI protocol), but \textbf{no framework we surveyed ships typed
structured delivery as the default end-to-end rendering path}, and the
public motivation is rendering predictability, never correctness
\citep{copilotkit2026genui}.

\paragraph{Evaluation methodology.} Turns within a dialogue are not
independent and turn-level significance claims frequently fail
cluster-robust correction \citep{schessl2026autocorrelation}; every gap interval here resamples \emph{dialogues}, not turns. The full
survey, including the structured-output infrastructure literature and the
closest dialogue benchmark (BOULDER \citep{kartac2026boulder}: it computes over given records; we ask whether the set survives delivery), is in Appendix~\ref{app:extended}.

\section{System and Conditions}
\label{sec:system}

\subsection{The deployed system}
\label{sec:deployed}

The system under study, AwardWise, is a
multi-agent travel concierge on Google's ADK: a planning
agent routes requests; sub-agents wrap the award (Seats.aero) and cash
(SerpAPI) search pipelines, which write the canonicalised record set to session state. Two
properties matter here.

\begin{enumerate}
\item \textbf{Search results do not enter the model's context.} The
  sub-agent's function response carries a count and the search
  parameters; records go to session state, and the client receives rows
  as typed events alongside, not through, the model's token stream.
\item \textbf{The model translates each follow-up's intent into
  arguments; a deterministic tool evaluates them.}
  The context tool evaluates filter arguments (vocabulary in C.10)
  over the stored record set. The remaining \emph{translation} job (the full language-understanding problem of tool use, scope anaphora
  pinned by convention, \S\ref{sec:authoring}) is where BYPASS's
  residual failures live (\S\ref{sec:failure}). The architecture changes
  the failure surface, not the difficulty: a mistranslated argument
  yields one wrong-but-real subset at any size, where re-emission puts
  every field of every row at stake on every turn.
\end{enumerate}

We name the conditions for what happens to the records: in
\textbf{SYNTHESIS} the record set passes through the model's generation
and is re-emitted; in \textbf{BYPASS} records reach the interface as
typed data and are never generated. The distinction is the delivery;
model, sampling and instruction lineage are held constant. BYPASS is this system
unmodified, driven over frozen payloads: its single model-side error
source is translation, and records are never re-typed, so field values
cannot mutate in transit.

\subsection{SYNTHESIS: the counterfactual fork-point architecture}
\label{sec:synthesis}

The system is a fork of the \texttt{adk-samples} travel-concierge
\citep{adksamples2025}, whose contract is the canonical ADK pattern: a
flight search sub-agent returns a full structured row list into the
planner's context, there is no context tool, and follow-ups are processed
in-head. SYNTHESIS rebuilds that architecture inside the mature product,
so delivery is the only difference: the production tree deep-copied with
exactly four deltas: a lineage-matched instruction variant, context tools removed,
records injected \emph{mechanically} into each sub-agent's summary so
SYNTHESIS is not penalised for sub-agent transcription noise, and output
schemas adjusted so the injected records survive (all four in Appendix~\ref{app:authoring}). The planner emits each turn's table
through an output schema (\texttt{flights}, \texttt{summary}), parsed
structurally rather than scraped from markdown; everything else (model,
sampling, other sub-agents, tools, state writes) is identical to
production.

Three consequences (in full in Appendix~\ref{app:authoring}):
\textbf{translation error is common-mode}: the same planner with the
same instruction lineage decides what the user asked for in both
conditions, so the residual gap is attributable to delivery;
\textbf{mechanical delivery makes SYNTHESIS stronger than its fork-true
form}, which passes records through an LLM emission before the planner;
and \textbf{BYPASS's advantage is joint}: synthesis bypass plus
structured-state preservation, not separated here
(Table~\ref{tab:components}; \hyperref[sec:limitations]{Limitations}).
Both conditions run Gemini 2.5 Flash at temperature 0.1 / top\_p 0.5,
inherited by lineage rather than chosen, so temperature cannot explain
the gap.

\begin{table}[t]
\centering
\small
\begin{tabular}{@{}p{0.26\linewidth}p{0.3\linewidth}p{0.3\linewidth}@{}}
\toprule
Component & SYNTHESIS & BYPASS \\
\midrule
Translate (NL $\to$ filter args) & model \emph{(identical)} & model \emph{(identical)} \\
Record storage & model context, re-emitted every turn & session state, written once \\
Follow-up filter & in-context, by the model & deterministic evaluator \\
Delivery to UI & parsed from model output & typed side channel \\
\bottomrule
\end{tabular}
\caption{What differs between the conditions. The headline gap measures
the joint effect of the last three rows; per-factor attribution requires
unbundled conditions we leave to follow-up work.}
\label{tab:components}
\end{table}

\section{Method}
\label{sec:method}

A note on instrument versions: partway through the study our measurement surfaced defects in production and in the instrument itself (\S\ref{sec:integrity}); after
fixing them we declared an \textbf{instrument freeze} and re-ran every
experiment. We label runs before those fixes \textbf{pre-fix} and runs on the frozen instrument \textbf{post-fix}; every headline number is post-fix. Pre-fix numbers appear only in the labelled
instrument-effect accounting (\S\ref{sec:whatmoved}), the attribution argument
(\S\ref{sec:ex1}), and the appendices.

\subsection{Freeze real, then replay}

The study is a \emph{freeze-real-then-replay} design: real searches ran
through the \textbf{production search tools}, not raw vendor APIs (hub routing, deduplication and canonicalisation are part of the system under study), and everything they returned was frozen. The frozen corpus is
\textbf{408 captures over 36 routes} holding \textbf{9,809 records}
(largest single payload 510 rows), spanning four date bands and two
cabins per route (composition and provenance in
Appendix~\ref{app:authoring}). Experiments serve the vendor boundary from
these captures, never a live vendor call.

\subsection{Dialogue authoring}
\label{sec:authoring}

Dialogue templates derive from demonstration sessions conducted by an
author, an experienced award-travel practitioner, over several rounds exercising representative award-hunting workflows. Observed
phrasings form the allowed vocabulary, and the 24 dialogues vary along
four knobs; the knobs, six dialogue arcs and turn families are inventoried in Appendix~\ref{app:authoring}.

Two authoring rules keep turns scoreable rather than arguable: \textbf{no
heuristic constraints} (``red-eye'' is banned; the mapping is pinned or
phrased explicitly, ``departing after 9pm'') and \textbf{no
undefined-aggregate superlatives} (a superlative must name a total order;
Appendix~\ref{app:authoring} works the example); the second was adopted
mid-study, retiring such a turn shipped in five dialogues
(\S\ref{sec:integrity}). Thresholds and boundary probes are pinned so
competing readings cannot both be defensible
(Appendix~\ref{app:authoring}).

A third convention pins down \textbf{scope semantics}, since every turn's
ground truth depends on what a follow-up operates over: a new filter
\textbf{composes} with the active scope (``only morning departures''
narrows the current table); a refer-back \textbf{restores} the view where
its referent was introduced (``show all the morning ones again'' returns
to the table that named ``the morning ones'', dropping filters and limits
applied after it); and a constraint revision \textbf{re-executes} the
turn that set the constraint against that turn's own base (``actually, up to 87,499 points is fine'' re-derives the cap from the full table, not the
narrowed one; the deliberately odd cap is a pinned boundary probe, one
point below the payload's real 87,500 value). A turn meaning to reset
must name its base explicitly
(``across ALL results\ldots'').

\subsection{The oracle}
\label{sec:oracle}

Per-turn ground truth is \textbf{computed, not annotated}: each
dialogue's frozen captures are replayed through the production search
tools, leaving exactly the session state a real search leaves, and each
filter turn's pinned arguments are passed to the production context tool's evaluator over that state; the returned envelope (the tool's function response: count and arguments, rows by reference) is the ground truth, normalised by the production
canonicalisation layer at a fixed SHA, applied identically to both
conditions. A liveness verifier rejects empty or trivially-repeated
ground truth on every turn before any model is called
(\S\ref{sec:integrity}; the verifier and the oracle's by-reference read
path in Appendix~\ref{app:authoring}).

\subsection{Metric}
\label{sec:metric}

The primary metric is \textbf{per-turn field fidelity with strict
count-mismatch zero-out}, the policy adopted from the single-turn study
\citep{hoang2026structured}; field-level scoring against source records
descends from data-to-text evaluation \citep{wiseman2017data2doc}.\footnote{A
post-submission instrument audit found that this metric can score a turn
as perfect when the emitted set has the right count but the wrong
membership, since field fidelity is computed over aligned pairs only.
Rescoring every turn with an exact-set rule (zero on any spurious or
missing row) leaves BYPASS essentially unchanged and lowers SYNTHESIS
slightly, so the reported gap is, if anything, understated.}
Emitted records are aligned with ground-truth records on the identity key
(\texttt{flight\_number}, \texttt{date}, \texttt{source}); scored fields
are those present and non-empty in the ground-truth record, drawn from a
fixed fifteen-field set; comparison is exact after production
canonicalisation (field list and key rationale in
Appendix~\ref{app:authoring}). When the emitted count equals the
ground-truth count, turn fidelity is the fraction of scored fields, over
all aligned record pairs, whose emitted value matches the ground truth.
If the counts differ, the turn's fidelity is instead \textbf{zero},
regardless of how many individual fields match. This is strict on purpose since the harm is downstream, i.e., the user acts on the table as shown. A silently missing award flight is a missed booking the user never
learns existed.

A turn counts as \textbf{correct} iff it produced an artifact at all and
its zero-out fidelity is 1.0. Turns where the deployed condition answered
from the count-only envelope without materialising rows currently count
as failures, so \textbf{BYPASS's published rate is a lower bound}.
Partial-credit fidelity (the field-match fraction without the zero-out: \textbf{98.2\% for BYPASS vs 82.7\% for SYNTHESIS}) and the ordered-turn axis (order-exact: \textbf{94.2\% vs 51.8\%}) are computed alongside every turn and reported in full in
Appendix~\ref{app:authoring}. Search turns establish context and are not
scored; award-versus-cash comparison turns exercise a different
production pipeline and are out of scope.

\subsection{Runs, repeats, and the analysis layer}
\label{sec:analysis}\label{sec:verdicts}

Each dialogue is run $k{=}3$ times per condition at production sampling;
repeats are independent, not seeded replays, so results reproduce
\textbf{distributionally, not turn-for-turn} (\hyperref[sec:limitations]{Limitations}). All published
aggregates load through a single corpus module that decides what counts:
network errors (429/503) are holes a recovery re-run may fill (a 429 is not data), and a run with no scoreable output scores zeros on its expected filter turns
(inclusion rules, per-run trace contents and the analysis-layer inventory in
Appendix~\ref{app:authoring}).

\paragraph{Hypotheses.} \textbf{H1a}: the fidelity gap appears at the
first filter turn. \textbf{H1b}: synthesis failures are not independent
across turns; history predicts failure. \textbf{H1c}: synthesis fidelity
degrades with the size of the delivered record set. A turn-count
\textbf{super-linearity} conjecture (failures compound faster with depth
than an independence null) was pre-registered but unsupported, published as
a diagnosed negative result (\S\ref{sec:depth}). The size instruments (\S\ref{sec:size}, \S\ref{sec:family}, \S\ref{sec:factorial}) were designed after the main corpus's first-turn rates were observed, each pre-specified (cells, byte-identical scored turns, expected ground-truth counts) before its runs. A second family, \textbf{H2} (demonstration-versus-instruction), was stated as motivation and not measured (\hyperref[sec:limitations]{Limitations}; Appendix~\ref{app:h2pilot}). Verdicts:
\textbf{H1c supported} as a controlled manipulation (\S\ref{sec:size});
\textbf{H1a supported} (\S\ref{sec:corpus}); \textbf{H1b supported}
(\S\ref{sec:cluster}).

\paragraph{Scale.} 408 captures / 9,809 records over 36 routes; 24
dialogues $\times$ $k{=}3$ runs $\times$ 2 conditions; 339 scored turns per condition carry the headline, on one
model (Gemini 2.5 Flash; scale table: Table~\ref{tab:scale}, Appendix~\ref{app:authoring}), with further families in \S\ref{sec:family} (Table~\ref{tab:family}, Appendix~\ref{app:results}).

\section{Results}
\label{sec:results}

\subsection{Delivered table size drives the synthesis collapse}
\label{sec:size}

The central result is a controlled manipulation: we vary only the size of
the table delivered into the agent's context, holding route, cabin, filter
vocabulary, turn wording and turn position fixed, and score the same
filter turn.

\begin{figure}[t]
  \centering
  \includegraphics[width=\columnwidth]{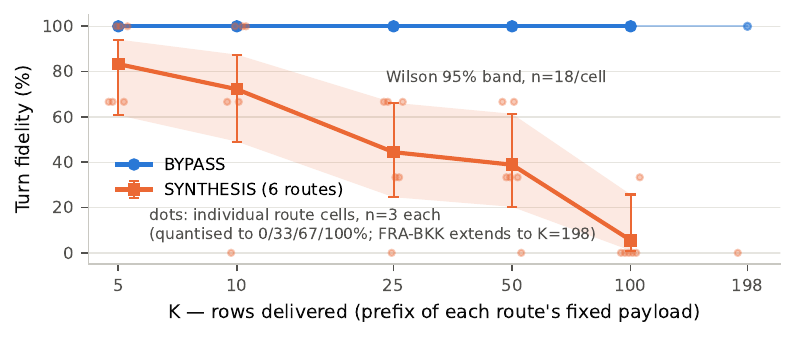}
  \caption{Turn fidelity vs K: the delivered table is the first K rows of each route's fixed payload, so size varies only by appending rows (six routes pooled). Light dots are per-route points at each K ($n{=}3$ runs each, hence quantised to 0/33/67/100\%); the band is a Wilson 95\% interval on the six routes pooled. Flash SYNTHESIS decays monotonically to 6\% at K=100;
  BYPASS is 100\% in all 93 runs.}
  \label{fig:topk}
\end{figure}

\paragraph{Content-controlled (top-K), replicated over six routes.} From
each of six routes' fixed payloads (112--198 rows) we deliver the first K rows: a true prefix, so the small table is literally the opening of the
large one and only the count differs. The same query (``only afternoon
departures'') is asked at every K; each cell pools 6 routes $\times$ 3
repeats ($n{=}18$; per-cell table: Table~\ref{tab:topk}, Appendix~\ref{app:sizedialogues}).

With only K varying, pooled Flash SYNTHESIS decays monotonically
from 83\% at K=5 to 6\% at K=100 while BYPASS is perfect in all 93 runs
(Figure~\ref{fig:topk}); FRA-BKK's full 198-row payload, the one route
run at K=198, also scores 0/3. The decay is graded, not a cliff: single routes at $n{=}3$ per cell quantise to 0/33/67/100\%, but pooled
over six routes the curve is a slope, and a Cochran--Armitage trend test
with log-K scores rejects a flat profile ($z = -5.07$, two-sided
$p = 3.9{\times}10^{-7}$; cluster-bootstrap robustness in
Appendix~\ref{app:results}; FRA-BKK failure anatomy in
Appendix~\ref{app:sizedialogues}). No change of route, date, price
distribution, or query is needed. Table size alone produces it.

\paragraph{Cross-route replication (date-varied).} Independently, on four
routes we vary delivered size by capture date; each route is a
single-cabin ladder (Figure~\ref{fig:ladder} and Table~\ref{tab:routes}, Appendix~\ref{app:sizedialogues}).

SYNTHESIS falls from 75--100\% on the smallest tables to near-zero beyond ${\sim}74$ rows in every route that reaches that size (the sole exception: FRA-BKK's 8\% at
194 against 0\% at 74, one turn in twelve); BYPASS is flat at
\textbf{99.2\% (238/240)}, its two misses a single starved run (the search sub-agent never fired; \S\ref{sec:starvation}). A control leg, a byte-identical 16-row table
in all twenty dialogues, scores SYNTHESIS 77.5\% and BYPASS 100\%, and holds even inside dialogues whose treatment turn scored zero (Appendix~\ref{app:sizedialogues}): the failure tracks the
table delivered at that turn, not the depth of the dialogue.

The date-varied ladder confounds size with content; the top-K ladder does
not, and they agree: the effect survives under a pure prefix manipulation. A size-by-position Latin square replicates the collapse at every position with no detectable position effect (\S\ref{sec:factorial}).

\subsection{The gap at corpus scale}
\label{sec:corpus}

Over 24 authored dialogues, three repeats: \textbf{BYPASS 92.3\%
(313/339), SYNTHESIS 45.7\% (155/339), gap +46.6pp} (dialogue-level
bootstrap CI $[+35.7, +57.1]$).
The gap is present at the first filter turn (not an accumulation effect) and never smaller than the per-route size ladders imply. The
direction is near-unanimous at the dialogue level: BYPASS beats SYNTHESIS
in 22 of 24 paired dialogues (one tie, one loss; exact sign test
$p{=}5.7{\times}10^{-6}$), in 24 of 24 on GPT-4o-mini and 21 of 24 on
Claude Haiku (\S\ref{sec:family}): 67 wins against 2 losses across the
three families.

\subsection{The gap is not specific to one model family, and a tier upgrade only rents it back}
\label{sec:family}

With both conditions re-run over the \textbf{full 24-dialogue corpus} on two further families with the same agent stack (sampling caveat: \hyperref[sec:limitations]{Limitations}; $n{=}339$ scored turns per cell, the headline's corpus),
the paired gap holds across all three families: \textbf{+46.6pp} [+35.7,
+57.1] (Gemini 2.5 Flash), \textbf{+49.6pp} [+39.2, +60.2] (GPT-4o-mini),
\textbf{+55.8pp} [+41.1, +69.9] (Claude Haiku 4.5); per-condition cells
with Wilson 95\% CIs are in Appendix~\ref{app:results} (Table~\ref{tab:family}). The gap converges: +46.6 to +55.8pp across
three families, every family's condition intervals disjoint. Family
determines \emph{who fails how}, not \emph{how much the architecture
costs}; BYPASS's floor (67.0--92.3\% across the three families) tracks translation quality (Table~\ref{tab:xfam-translation}), yet no family's BYPASS falls below any of the three families' SYNTHESIS; the exception is the Pro tier (72.0\%, at 12$\times$ BYPASS-on-Flash's price).

The failure shapes are family-specific. \textbf{Gemini corrupts}: real
rows, wrong set (\S\ref{sec:failure}). \textbf{GPT garbles}: unparseable
JSON mid-table, type-broken fields, dropped required fields.
\textbf{Claude omits}, concentrated on larger tables: well-formed JSON that drops the required \texttt{flights} array while the prose summary counts the delivered table correctly (an accurate count attached to a missing table). One mechanism, three masks
(specifics, per-family latencies, and a small-table pilot artifact in
Appendix~\ref{app:results}).

The taxonomy has a load axis, and it is the paper's thesis in one picture. Re-run on all four models with identical K-row prefixes of the same frozen 198-row payload, the content-controlled ladder (\S\ref{sec:size}) shows the \textbf{collapse threshold ordering by model capability}: under 10 rows for GPT-4o-mini, 10--25 for
Claude Haiku, 50--100 for Gemini Flash, while Gemini Pro alone degrades gracefully (perfect through 25 rows, 83\% at 50, 67\% at 100, and
50\% at the full 198), and BYPASS is size-flat for all four models
(140/144 across the ladder; Figure~\ref{fig:family}). Capability buys
threshold; architecture buys size-immunity. Cells pool six repeats.

A same-family \textbf{tier} upgrade only rents the gap back: over the full corpus ($n{=}339$ per cell) Gemini 2.5 Pro halves the gap rather than closing it (BYPASS 92.6\% vs SYNTHESIS 72.0\%, +20.6pp [+11.7, +31.4] against Flash's +46.6pp) at 5$\times$ Flash's per-turn cost and 1.6$\times$ its latency. BYPASS is tier-insensitive (92.3\% Flash, 92.6\% Pro), and BYPASS on the cheap production Flash beats SYNTHESIS on Pro at one twelfth the price (detail in Appendix~\ref{app:results}). Tier is the top rung of
Figure~\ref{fig:family}'s ordering: it raises the collapse threshold; it
does not remove the mechanism.

\subsection{What failure looks like}
\label{sec:failure}

Post-fix, SYNTHESIS no longer fabricates on delivered data
(\S\ref{sec:qualitative}): of 99 failed SYNTHESIS turns in the size
ladder, 96 have partial-credit fidelity ${\approx}1.0$: \textbf{the rows emitted are real, but the set boundary is wrong}. The dominant mode
is truncation with a cascade: a turn truncates a 74-row result to 55
emitted, and the next turn filters \emph{that} emission rather than the
true set, the pre-fix mechanism (\S\ref{sec:qualitative}) on real
rows. BYPASS cannot exhibit either mode; its failures are wrong-but-real
filter arguments: of its 26 failed turns, 22 applied wrong arguments,
3 applied the right arguments over an under-delivered result, and one
produced no artifact (anatomy in Appendix~\ref{app:results}).

\subsection{Cost and latency, not only accuracy}
\label{sec:cost}

\begin{figure}[t]
  \centering
  \includegraphics[width=\columnwidth]{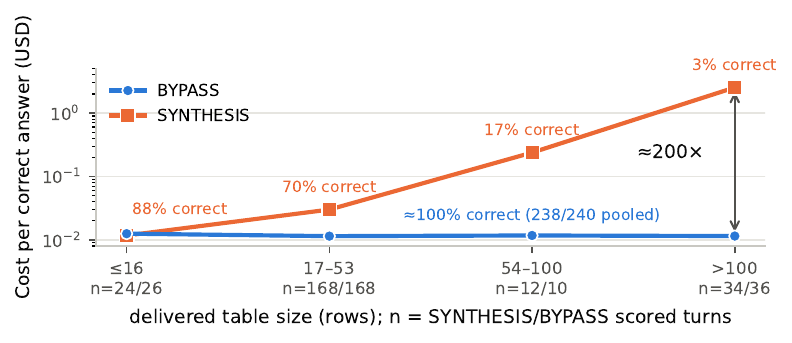}
  \caption{Cost per correct answer by delivered-size band (size-ladder sweeps; mean turn cost $\div$ accuracy, log scale). BYPASS is flat at ${\approx}\$0.012$; SYNTHESIS climbs to \$2.53 beyond 100 rows, ${\approx}200\times$. Point labels give SYNTHESIS's per-band accuracy (one correct SYNTHESIS turn in the ${>}100$ band; the multiple is an order-of-magnitude estimate). Unequal $n$: the starved ORD-NRT run's two BYPASS turns (0 of 72 rows delivered) count in ${\le}16$, its only misses; two ${>}100$-row SYNTHESIS turns emitted nothing scoreable (Appendix~\ref{app:sizedialogues}).}
  \label{fig:cpc}
\end{figure}

On the full 24-dialogue grid, mean per-turn latency is \textbf{BYPASS
4.9s vs SYNTHESIS 31.7s}, a 6.5$\times$ penalty. BYPASS's latency and cost are flat in table
size because it re-emits nothing, while SYNTHESIS's scale with the set it
must reproduce (Figure~\ref{fig:cost}, Appendix~\ref{app:results}). At the largest tables a
synthesis turn averages 146 seconds (194 at worst) and costs 8.7$\times$ BYPASS's flat per-turn cost, for an 8\%-fidelity answer. Dividing cost by accuracy (Figure~\ref{fig:cpc}):
BYPASS produces a \emph{correct} answer for ${\approx}\$0.012$ at every
table size, while SYNTHESIS's cost per correct answer climbs to \textbf{\$2.53 beyond 100 rows, roughly 200$\times$}; tables that size are precisely the regime a search product exists for. Model tier does not escape this arithmetic (\S\ref{sec:family};
Appendix~\ref{app:results}): a model upgrade rents part of what the
architecture gives outright, and the rent is high.

\subsection{Errors cluster, only synthesis propagates values, and the depth test finds no super-linearity}
\label{sec:cluster}\label{sec:depth}

Two independent instruments show synthesis failures are not independent
draws, and bypass failures are. Lag-1 conditioning: after a failed turn,
SYNTHESIS fails again 71.9\% of the time against 31.4\% after a clean
turn (a \textbf{recovery rate of 28.1\%}), while BYPASS recovers
\textbf{95.7\%}. Value-tracing (a deliberately liberal upper bound; C.4): \textbf{53 of SYNTHESIS's 177 traceable
failures (30\%) are propagated; 0 of BYPASS's 25 are}: for BYPASS no model-emitted set exists to trace, so propagation is precluded and the zero reflects that no counterexample arose. The mechanism is architectural:
BYPASS recomputes each turn from preserved state; SYNTHESIS filters its
own poisoned output, and its residual failures are isolated translation
slips.

Our pre-registered turn-count super-linearity test finds none. The reason
is structural: with SYNTHESIS's 56.9\% first-turn failure rate the accumulation-only null saturates, leaving almost no measurable headroom above it; the test is degenerate in its own high-error regime (construction,
excess-by-position and sample sizes in Appendix~\ref{app:authoring}).
This redirected the analysis to table size (\S\ref{sec:size}), which is
monotone, controlled, and does not saturate.

\section{Qualitative failure analysis}
\label{sec:qualitative}

A grounding census classifies every emitting SYNTHESIS turn under both instrument versions (production model, 1,306 turns) by grounding: is each emitted flight
number present in anything ever delivered into the run? The result is
categorical: \textbf{fabrication tracks grounding absence; corruption
tracks re-emission.} Under neither version did the model invent a table when the
queried table had been delivered; every from-nothing fabrication sits
over a missing one, and on fully empty context (31 turns, each downstream
of a starved search) fabricated tables average 1.4--2.1 rows. A tiny
emission is thus a \emph{signature of absence} (full decomposition in
Appendix~\ref{app:census}).

\subsection{After the prompt fix: the fabrication is gone, the corruption
is not}
\label{sec:postfix}

On the frozen instrument we audited every failed SYNTHESIS turn in the
size ladder, 99 of them. \textbf{96 of the 99 carry partial-credit
fidelity ${\approx}1.0$: every row the model emits is real.} Fabricated
tables were never a data-present phenomenon; post-fix, fabrication
persists exactly where residual starvation persists (19 empty-context
turns, mean 1.4 rows) and nowhere else. What did not vanish is the set
error. The dominant modes:

\begin{itemize}
\item \textbf{Truncation.} Asked to show a 74-row result, the model emits
  55 of the 74 (real rows, correctly formatted, silently incomplete;
  \texttt{d26\_fra\_bkk\_size\_m\_early} t2: GT 74, emitted 55, partial
  1.0).
\item \textbf{Cascade on its own emission.} The next turn then filters
  the emitted 55 rather than the true 74: asked for the afternoon subset
  (GT 11), it returns 45: real rows over a corrupted base and a corrupted scope (same dialogue, t3). This is Exhibit 1's propagation mechanism (Appendix~\ref{app:census}), on real rows instead of invented ones.
\item \textbf{Identifier splicing under load.} In 25 of 943 post-fix populated emitting turns an identifier is corrupted inside an otherwise-real row set, concentrated in large-table runs (specifics in D.1): field-level corruption that count-based checks cannot see.
\end{itemize}

\subsection{The pre-fix runs pin the attribution (Exhibits 1--2,
Appendix~\ref{app:census})}
\label{sec:ex1}

The obvious deflationary reading, that synthesis merely had a bad
prompt, is answered by comparing runs before and after the prompt repair.
Pre-fix, the production prompt itself taught fabrication
(\S\ref{sec:integrity}): in one starved run SYNTHESIS answered a points
filter over an undelivered payload with a plausible schema-valid table,
then filtered its own fiction for five more turns; constrained decoding guaranteed the shape of the lie. BYPASS's failure on the same starvation was visible emptiness: its table comes from state, and empty state renders empty (turn by turn in Appendix~\ref{app:census}, Tables~\ref{tab:ex1}--\ref{tab:ex2}). Repairing the prompt \textbf{removed the
fabrication but not the failure}: fabrication tracked grounding absence
under both prompts, while set corruption survived every prompt repair,
scales with the delivered table (\S\ref{sec:size}), and BYPASS cannot exhibit it. Attributing the gap to bad prompting is backwards: the better the prompt got, the more purely the
residual gap measured the architecture.

\section{Measurement integrity}
\label{sec:integrity}

\subsection{We found three defects in our own ground truth, and report
them}
\label{sec:gtdefects}

The range/flex arc initially scored \textbf{BYPASS 14/27 (52\%) vs
SYNTHESIS 16/27 (59\%)}, the only arc synthesis won. Challenged rather
than published, it dissolved into three instrument defects (full
narratives in Appendix~\ref{app:integrity}); after correction:
\textbf{BYPASS 25/25 (100\%), SYNTHESIS 12/25 (48\%)}, a 59-point swing (from $-7$pp to $+52$pp; turn accounting in E.1). The \S\ref{sec:oracle} liveness verifier
exists because of this, the corrected arc replaces the defective one in
\emph{both} conditions, and the same arc captured a user-facing production bug, fixed (Appendix~\ref{app:integrity}).

\subsection{The starvation event: diagnosed and largely removed}
\label{sec:starvation}

In pre-fix runs, the search sub-agent sometimes never fired despite a ready verdict, failing the dialogue wholesale (21.2\% of SYNTHESIS
search turns, 5.2\% of BYPASS's). The mechanisms differ by condition
(Appendix~\ref{app:integrity}); repairs took the post-fix rates on the main 24-dialogue corpus to \textbf{0.0\% (BYPASS) and 9.7\% (SYNTHESIS)} (starved runs persist on other legs; \S\ref{sec:size}), and affected turns are
reported as failures, not excluded. Excluding starved runs \emph{widens} the pre-fix gap (re-scored: $+35.3 \rightarrow +37.5$pp): the confound suppressed the effect rather than manufacturing it. Post-fix, excluding the 11 of 72 starved SYNTHESIS runs (BYPASS loses none)
lifts SYNTHESIS to 47.4\% (gap +44.9pp): the headline does not rest on residual starvation.

\subsection{Instrument fixes: one move unexplained}
\label{sec:whatmoved}

Fixing the instrument moved the headline on the comparable subset (the
19 dialogues whose turn sets are identical under both instrument
versions; 264 turns per condition) from $+33.7$pp to $+49.6$pp. The BYPASS side is accounted for (removing starvation and the
fabrication-teaching prompt lifted it from 81.4\% to 91.3\%); the
SYNTHESIS side is not: two fixes expected to \emph{raise} it by roughly
4pp preceded a fall from 47.7\% to 41.7\%, a $-6.0$pp residual exceeding the pre-fix corpus's own sweep-to-sweep (repeat-to-repeat) spread, which we flag as unexplained rather than absorb (full accounting in
Appendix~\ref{app:integrity}). We declared the freeze before the final
runs and report both readings; under either instrument version the gap
exceeds $+33$pp.

\subsection{A factorial separates size from position}
\label{sec:factorial}

In the 24-dialogue corpus, table size and turn position are correlated,
so observational stratification alone cannot separate a size account from a first-turn account. The size ladders (\S\ref{sec:size})
counterbalance position, and a \textbf{3$\times$3 size-by-position Latin
square} settles it: size at fixed position (slot) is decisive
(92$\rightarrow$33$\rightarrow$0\% at slot 1; $\chi^2{=}39.4$,
$p{=}3\times10^{-10}$, surviving cluster bootstrap in 2000 of 2000
resamples) while no position main effect is detectable ($\chi^2{=}0.8$, $p{=}0.36$);
BYPASS crosses the same square at 106/108. Regression, design and the secondary texture are in Appendix~\ref{app:integrity}.

\section{Conclusion}
\label{sec:conclusion}

In a deployed multi-turn agent, re-emitting a tool's records through the model's context each turn costs 46.6pp of per-turn field fidelity against carrying them as state on a typed side channel. The load variable is the \textbf{size of the re-emitted record set}: synthesis fidelity falls from 75--100\% to ${\le}8\%$ beyond ${\sim}74$ rows, while BYPASS holds 99.2\% at flat latency and cost. The gap appears at the first filter turn, the hazard is flat with depth, and our pre-registered
super-linearity test finds none: its null saturates in the high-error regime it targeted.
Synthesis failures are not independent (\S\ref{sec:cluster}); BYPASS structurally cannot inherit one: it never re-emits records. Practitioners adopted typed side-channel delivery for rendering; this paper supplies the joined measurement from one deployed system,
suggesting the field's UX-motivated convergence was also, unknowingly, a
correctness decision.

\section*{Limitations}
\label{sec:limitations}

\paragraph{Hypothesis H2 is motivated, not measured.} Earlier engineering
observations suggested that an answer-shaped demonstration in context
overrides an on-point prompt prohibition (Appendix~\ref{app:h2pilot}
reproduces the pilot that motivates this); the pre-registered $2\times2$
design of seed shape $\times$ prompt strength remains future work, and
this paper asserts nothing about H2.

\paragraph{The size manipulation has bounded coverage.} The
content-controlled ladder covers six routes but one cabin, with per-route
ceilings of 112--198 rows against the 510-row payloads we hold; we
declined a true ${\sim}200$+ band through range captures because that
would exchange the position confound for a query-shape confound at
exactly the cell carrying the effect. Whether the decay moves with route
complexity, cabin or record width is open, and the cell resolution
localises the shape only coarsely: the pooled decay is monotone from
83\% at K=5 to 6\% at K=100, with no single knee we can name.

\paragraph{Headline numbers from one family; no seeds, no bit-reproducibility.} The headline number is Gemini 2.5 Flash's; \S\ref{sec:family}'s family and tier legs replicate the gap at the same scale, so the residual limitation is reproducibility, not coverage. Runs are independent repeats at production sampling
(temperature 0.1 / top\_p 0.5) with \textbf{no random seed set}, on a
non-bit-deterministic serving stack: reproducible distributionally, not
turn-for-turn; payload identity is derivable from dialogue specifications
and capture manifests, but no content hash is in the traces. Cross-family
sampling parity is imperfect: Anthropic's API rejects temperature and
top\_p together, so Claude runs clear top\_p;
Appendix~\ref{app:parity} reports the measured direction of that deviation (a Gemini-only ablation, the one family where the combination is settable): if the direction transfers across families, the parity break biases against BYPASS and the Claude gap is conservative.

\paragraph{The comparison is joint by construction.} As stated in
\S\ref{sec:synthesis}, BYPASS combines synthesis bypass with
structured-state preservation and we do not separate them. A natural
further condition would keep the same cross-turn state mechanism but
deliver through the LLM; we did not run it, because it corresponds to no
deployed pattern we are aware of. The cost is that our effect size is
attributable to the pair, not to either member. One partial unbundling
comes free: every scored turn of the content-controlled ladder
(\S\ref{sec:size}) is its run's first filter turn, so no prior emission
exists for state preservation to protect, and the collapse appears
anyway; the synthesis half of the pair suffices to produce the size
effect.

\paragraph{Statistical resolution.} CIs are wide (dialogue-level
bootstrap: BYPASS 88.3--95.9, SYNTHESIS 34.8--57.2; gap [+35.7, +57.1])
because $n{=}24$ dialogues is the resampling unit, and strata are small:
the largest-size $\times$ deepest-position stratum holds 7 turns, and the two deepest turn positions are reached by only 8 and 3 dialogues respectively.

\paragraph{External validity and scope.} Captures are restricted to
non-stop and one-stop itineraries; turn scripts derive from expert
demonstration sessions, not production traffic; every payload is a
flight itinerary from one deployment; hotels and other verticals are out
of scope. Scope conventions are common-mode: the \S\ref{sec:authoring} rulebook
was fixed before runs and applied identically to both arms, so a
defective convention shifts both conditions' levels rather than the
between-arm contrast (the \S\ref{sec:gtdefects} corrections were applied
to both arms). Ground truth and the deployed arm share the production
evaluator, so a shared evaluator defect would depress SYNTHESIS alone;
the checks that do not share that code path are the authored
expected-row assertions (Appendix~\ref{app:sizedialogues}), the pre-run
liveness verifier (\S\ref{sec:oracle}), and the hand audits of all 99
ladder failures against the frozen payloads (\S\ref{sec:postfix}). An independent re-implementation of the filter semantics, checking the oracle from outside the production codebase, is the verification this design still lacks (independence detail: C.2). The
corpus-level marginal (+46.6pp) depends on the authored mix of
delivered-table sizes; the size-conditional curves
(Figures~\ref{fig:family} and~\ref{fig:topk}, Tables~\ref{tab:routes}
and~\ref{tab:topk}) are the transportable result, and readers can
reweight them under any deployment's size distribution. Vendor terms preclude releasing the payloads; the method
(\S\ref{sec:method}) replicates on any deployment's own data.

\paragraph{Other open items.} An outcome category for
count-only-envelope turns would raise BYPASS's measured rate honestly; a
greedy-decoding ablation and a temperature sweep were designed and not
run.

\section*{LLM and Agent Usage Disclosure}

\paragraph{Research system:} the system under study is itself a
multi-agent system (Google ADK, Gemini 2.5 Flash). Both conditions were
driven against production code; ground truth was computed from frozen
real vendor payloads (Seats.aero, SerpAPI), not model-generated.

\paragraph{Affiliation:} the authors are affiliated with AwardWise, the
company that operates the production system under study.

\paragraph{Experimental design and analysis:} the study design (the two
conditions, the size and prefix manipulations, the dialogue scripts, the
fidelity metric and its zero-out policy, and every ground-truth semantics
decision for scope composition, refer-back restoration, and constraint
revision) was specified and adjudicated by the authors, who personally
audited the exhibits, per-turn tables, and failure cases reported here.

\paragraph{Writing and code:} an AI assistant (Claude, Anthropic) was
used substantially throughout: under the authors' direction it drafted
and revised the paper's prose, implemented the replay harness and
analysis code, executed experiment runs, and performed reference
verification. Every quantitative claim traces to a committed artifact
produced by deterministic scripts. The authors reviewed the analyses,
directed each correction, and bear sole responsibility for all claims,
interpretations, and final wording.

\paragraph{Artifact availability:} the camera-ready artifact will
release the 24 dialogue templates with knob specifications and the
\S\ref{sec:authoring} scope-semantics rulebook; the replay harness,
including the four-delta SYNTHESIS construction (C.9) and the
model-override layer implementing sampling parity and the top\_p
ablation (E.5); the analysis layer, which reproduces every published
number from committed result files; capture manifests (producing git
SHA, request parameters, and payload content hashes); per-run traces
with vendor payload contents stripped or hashed (tool calls, arguments,
token usage, per-turn scores); and the internal pre-registration text
with its commit hashes. Vendor terms preclude releasing the frozen
payloads themselves.

\nocite{google2026a2ui}
\bibliography{references}

\appendix

\section{The Size-Manipulation Dialogues and Per-Route Ladders}
\label{app:sizedialogues}

The manipulation should be checkable rather than taken on trust, so the
dialogue set is reproduced here. The tables below describe the original
d25--d30 six-dialogue design, which sits within the larger 20-dialogue
ladder of \S\ref{sec:size}. Every dialogue has the same six-turn shape
(search, two scored filter turns, second search, two more scored filter
turns) with \textbf{turns 2, 3, 5 and 6 scored and byte-identical across all six dialogues} (Table~\ref{tab:script}); the only textual difference is the date in the two
search turns, the manipulation instrument, because the capture date determines how large the delivered table is (Table~\ref{tab:cells}). Each dialogue contains one
\textbf{treatment leg} (size varies) and one \textbf{control leg} (always the same \texttt{close\_in} capture: 25 delivered rows, 16 after the seat filter); the
\texttt{\_early} / \texttt{\_late} suffix counterbalances which leg comes first in the dialogue.

\begin{table*}[t]
\centering\small
\setlength{\tabcolsep}{4pt}
\begin{tabular}{lllll}
\toprule
dialogue & treatment size & leg order & first search date & second search date \\
\midrule
\texttt{d25\_fra\_bkk\_size\_s\_early} & 9 rows & treatment first & 2027-06-24 (release\_edge) & 2026-08-14 (close\_in) \\
\texttt{d26\_fra\_bkk\_size\_m\_early} & 74 rows & treatment first & 2026-10-14 (near) & 2026-08-14 (close\_in) \\
\texttt{d27\_fra\_bkk\_size\_l\_early} & 194 rows & treatment first & 2027-04-14 (far) & 2026-08-14 (close\_in) \\
\texttt{d28\_fra\_bkk\_size\_s\_late} & 9 rows & control first & 2026-08-14 (close\_in) & 2027-06-24 (release\_edge) \\
\texttt{d29\_fra\_bkk\_size\_m\_late} & 74 rows & control first & 2026-08-14 (close\_in) & 2026-10-14 (near) \\
\texttt{d30\_fra\_bkk\_size\_l\_late} & 194 rows & control first & 2026-08-14 (close\_in) & 2027-04-14 (far) \\
\bottomrule
\end{tabular}
\caption{The six size-manipulation cells. Sizes are \S\ref{sec:size}'s design labels (post-seat-filter counts); the delivered payloads are 17, 103 and 198 rows (control 25).}
\label{tab:cells}
\end{table*}

\begin{table*}[t]
\centering\small\setlength{\tabcolsep}{3.5pt}%
\begin{tabular}{clll}
\toprule
t & speaker turn & role & ground truth \\
\midrule
1 & ``Find award flights from FRA to BKK on 2027-06-24, economy class'' & search (establishes context) & --- \\
2 & ``At least 2 seats please'' & \textbf{scored} $\cdot$ fresh filter & 9 rows \\
3 & ``Only afternoon departures'' & \textbf{scored} $\cdot$ composes with previous & 3 rows \\
4 & ``Find award flights from FRA to BKK on 2026-08-14, economy class'' & search (establishes context) & --- \\
5 & ``Same here --- at least 2 seats'' & \textbf{scored} $\cdot$ fresh filter & 16 rows \\
6 & ``Afternoon departures only, please'' & \textbf{scored} $\cdot$ composes with previous & 5 rows \\
\bottomrule
\end{tabular}
\caption{The dialogue script, shown for \texttt{d25} (treatment first);
the other five differ only in the two dates; their ground-truth counts differ accordingly (Table~\ref{tab:gtcounts}).}
\label{tab:script}
\end{table*}

Turn 2/5 applies \texttt{min\_seats=2}; turn 3/6 adds the published
afternoon bucket (12:00--17:59) and composes with the previous turn.
Ground truth is computed by the production filter evaluator over the
frozen payload, never annotated (Table~\ref{tab:gtcounts}).

\begin{table}[t]
\centering\small
\begin{tabular}{lll}
\toprule
dialogue & treatment leg & control leg \\
\midrule
\texttt{d25} (S, early) & 9 $\rightarrow$ 3 & 16 $\rightarrow$ 5 \\
\texttt{d26} (M, early) & 74 $\rightarrow$ 11 & 16 $\rightarrow$ 5 \\
\texttt{d27} (L, early) & 194 $\rightarrow$ 87 & 16 $\rightarrow$ 5 \\
\texttt{d28} (S, late) & 9 $\rightarrow$ 3 & 16 $\rightarrow$ 5 \\
\texttt{d29} (M, late) & 74 $\rightarrow$ 11 & 16 $\rightarrow$ 5 \\
\texttt{d30} (L, late) & 194 $\rightarrow$ 87 & 16 $\rightarrow$ 5 \\
\bottomrule
\end{tabular}
\caption{Ground-truth row counts per cell.}
\label{tab:gtcounts}
\end{table}

The control leg is identical in all six dialogues (16 $\rightarrow$ 5
rows): synthesis runs whose 194-row treatment turn earned zero credit (one via an unscoreable emission) still average 90\% (9/10 control turns) on the byte-identical 16-row control leg of the same dialogues, which is what separates delivered-table load from context
accumulation. Every scored turn passed an oracle liveness gate before any
model was run (empty or trivially-repeated ground truth is rejected at
authoring time), and the authored \texttt{expected\_gt\_rows} is asserted
against the evaluator's \texttt{total\_matched} at run time, so payload
drift cannot silently rescale the treatment variable.

\subsection{The per-route date-varied ladders (\S\ref{sec:size})}

\begin{figure}[t]
  \centering
  \includegraphics[width=\columnwidth]{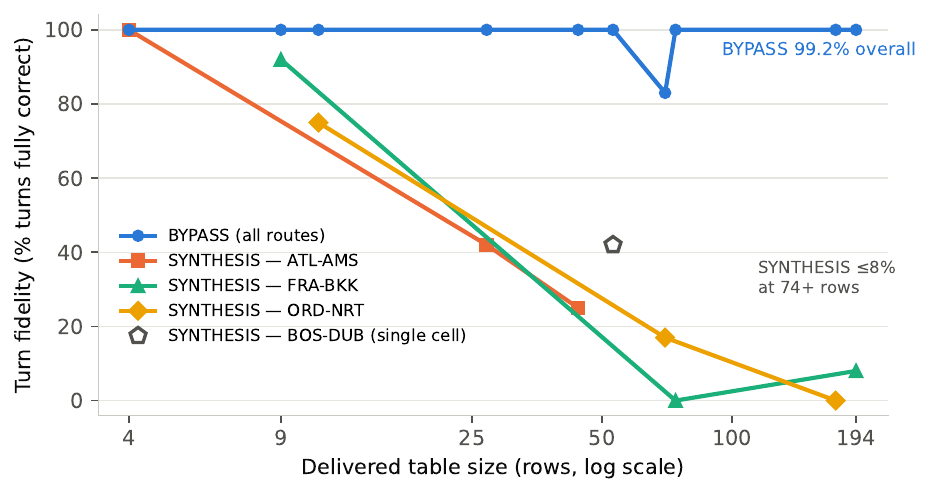}
  \caption{Turn fidelity vs table size. SYNTHESIS falls from 75--100\% on the smallest tables to $\le$8\% beyond 74 rows in every route that reaches that size (one 0\%$\rightarrow$8\% wobble at FRA-BKK's largest cell); BYPASS is 100\% everywhere except a single dip (70 rows, 83\%), one starved run (\S\ref{sec:integrity}); 99.2\% overall.}
  \label{fig:ladder}
\end{figure}

Each route is a single-route, single-cabin ladder varying delivered size
by capture date; Figure~\ref{fig:ladder} plots these cells
(Table~\ref{tab:routes}).

\begin{table}[t]
\centering\small
\setlength{\tabcolsep}{3pt}
\begin{tabular}{lll}
\toprule
route & rows $\rightarrow$ SYNTHESIS acc. & BYPASS \\
\midrule
FRA-BKK & 9$\,{\to}\,$92\% $\cdot$ 74$\,{\to}\,$0\% $\cdot$ 194$\,{\to}\,$8\% & 100\% \\
ORD-NRT & 11$\,{\to}\,$75\% $\cdot$ 70$\,{\to}\,$17\% $\cdot$ 174$\,{\to}\,$0\% & 100\%* \\
ATL-AMS & 4$\,{\to}\,$100\% $\cdot$ 27$\,{\to}\,$42\% $\cdot$ 44$\,{\to}\,$25\% & 100\% \\
BOS-DUB & 53$\,{\to}\,$42\% & 100\% \\
\bottomrule
\end{tabular}
\caption{Date-varied ladders. *ORD-NRT's BYPASS column excludes one
starved run (10/12 at the 70-row cell, the single dip visible in
Figure~\ref{fig:ladder}); with it, 99.2\% overall (238/240 scored turns). Figure~\ref{fig:cpc}'s $n$ row shares this accounting: the starved run's two turns count in the ${\le}16$ band at their delivered size of 0 rows, and two SYNTHESIS turns (d27 sweep k2 and d33 sweep k3, the 194- and 174-row cells; delivered payloads 198 and 190 rows) emitted nothing scoreable and drop from the ${>}100$ count.}
\label{tab:routes}
\end{table}

\subsection{The top-K ladder cells}

The five pooled cells of the content-controlled (top-K) ladder
(\S\ref{sec:size}); these are exactly Figure~\ref{fig:topk}'s plotted
points (Table~\ref{tab:topk}).

\paragraph{Failure anatomy on the content-controlled (top-K) ladder,
FRA-BKK.} Three failures were residual starvation (the search sub-agent reported success but its inner search never delivered rows, \S\ref{sec:starvation}; one in the full-payload K=198 cell, outside this table's K range), and the mid-K failures are genuine truncations whose self-reported counts exceed their emitted rows by one (claimed 12, emitted 11; claimed 19, emitted 18). The K=100 collapse is starvation-free: all three FRA-BKK runs received their 100 rows and truncated their answers (20/30/34 rows emitted against a ground truth of 42).

\begin{table}[t]
\centering\small
\setlength{\tabcolsep}{4pt}
\begin{tabular}{lccccc}
\toprule
K delivered & 5 & 10 & 25 & 50 & 100 \\
\midrule
SYNTHESIS & 83\% & 72\% & 44\% & 39\% & \textbf{6\%} \\
 & (15/18) & (13/18) & (8/18) & (7/18) & (1/18) \\
BYPASS & 100\% & 100\% & 100\% & 100\% & 100\% \\
 & (18/18) & (18/18) & (18/18) & (18/18) & (18/18) \\
\bottomrule
\end{tabular}
\caption{Content-controlled prefix ladder pooled over six routes. FRA-BKK's single-route full-payload cell (K{=}198, $n{=}3$: SYNTHESIS 0/3, BYPASS 3/3) is not pooled here; it completes \S\ref{sec:size}'s 93-run BYPASS total.}
\label{tab:topk}
\end{table}

\section{The Demonstration-versus-Instruction Pilot (H2)}
\label{app:h2pilot}

During earlier engineering work, seeding a session with an
\emph{answer-shaped} model turn (``Found 40 award options \ldots'' followed by a rendered table) made the agent pattern-complete the next
filter request without calling the filter tool, fabricating a count
(40$\rightarrow$10, 7$\rightarrow$2, 2$\rightarrow$1 across three
probes). A local A/B on one route, cabin and date (5 runs per variant):
an answer-shaped seed with an explicitly strengthened system-prompt
prohibition produced 0/5 grounded turns; a tool-trajectory seed produced
5/5 under either prompt. \textbf{$n$ is small, uncounterbalanced, one
route and one model: not a controlled experiment}, motivating only the
hypothesis that in-context demonstration governs grounding more strongly
than an on-point prompt prohibition. The pre-registered design, a
$2\times2$ of seed shape $\times$ prompt strength over ${\sim}30$
stratified seeded contexts $\times$ 5 runs per cell, is future work.

\section{Method and Analysis Detail}
\label{app:authoring}

\subsection{The undefined-aggregate rule, worked}

``Which day is cheapest?'', asked after a date-\emph{range} search whose
result spans several days, names no total order: ranking days requires a
day-level aggregate over per-flight prices (the day's minimum fare? its
mean? award points or cash?) that the utterance never picks, so several
answers are defensible and the turn is unscoreable by construction. The
\S\ref{sec:authoring} rule retired exactly such a turn, which had shipped
in five dialogues; its replacement is an unambiguous window-spanning
filter turn (``Morning departures across all the days'';
Appendix~\ref{app:integrity}).

Thresholds are authored as payload-pinned percentiles with a lint
asserting 20--80\% survival; boundary probes are pinned strictly between
two adjacent real values, so strict-versus-inclusive readings cannot both
be defensible.

\subsection{The oracle's read path}

Because the deployed system serves large result sets by reference (rows
in state, envelope carrying a pointer), the oracle reads the full served
row set through the same state path the client's stream uses, rather than
trusting the envelope's inline rows. The liveness verifier of
\S\ref{sec:oracle} asserts, before any model is called, that no turn's
ground truth is empty or a trivial repeat of the previous turn's, on
every turn of every dialogue. Three properties keep the oracle independent of the arm that shares its code. First, it executes the authored arguments, pinned in the dialogue spec, never the model's: BYPASS's own translations are scored against it and fail when the divergence changes the selected rows (22 wrong-argument failures; F.5), so the deployed arm is not correct by construction. Second, the authored \texttt{expected\_gt\_rows}, derived by hand from each frozen payload, is asserted against the evaluator's \texttt{total\_matched} on every scored turn of every dialogue, across all turn families (C.10). Third, the evaluator is challenged rather than trusted: row-level hand audits cover all 99 ladder failures and both exhibits (\S\ref{sec:postfix}, Appendix~\ref{app:census}), and the \S\ref{sec:gtdefects} challenge found three ground-truth defects, fixed with both arms re-scored. We use the production evaluator because it implements exactly what the oracle needs, a deterministic filter over frozen state under the \S\ref{sec:authoring} semantics; a second implementation built only for scoring would add disagreements needing their own adjudication without removing the shared-specification risk. An independent re-implementation of the \S\ref{sec:authoring} semantics is the check this design still lacks (\hyperref[sec:limitations]{Limitations}).

\subsection{Partial credit and order, in full}

Partial-credit fidelity over aligned pairs, post-fix: \textbf{98.2\%
for BYPASS and 82.7\% for SYNTHESIS} (mean over scored turns with an emitted artifact, $n{=}338$ and $n{=}333$; pre-fix: 93.3\% and 72.8\%), a 15.5-point gap against
the zero-out headline's 46.6.

Turns flagged \texttt{ordered} (limit-N superlatives and explicit re-sort
requests) are scored on emitted sequence as \textbf{order-exact} (the
emitted sequence equals the ground-truth sequence) and \textbf{order
agreement} (the fraction of positions holding the correct record). On the post-fix corpus's 86 (BYPASS) and 85 (SYNTHESIS) ordered turns:
order-exact 94.2\% vs 51.8\%, mean order agreement 95.9\% vs 58.2\%
(pre-fix: 93.7\% vs 61.2\% and 94.2\% vs 64.0\%).

\subsection{The analysis layer's instruments}

The corpus module's inclusion rules: three independent sweeps per
condition (the $k{=}3$ repeats); the range/flex arc taken only from its corrected re-run (\S\ref{sec:gtdefects}), the defective original excluded in \emph{both} conditions; recovery re-runs may fill a dialogue lost to a network error; a run producing no scoreable output is scored as zeros on its expected filter turns.

Beyond the aggregates named in \S\ref{sec:analysis}, the analysis layer
computes per-position marginal hazard; cumulative failure versus the
accumulation-only null $1-(1-p)^n$; and a per-dialogue
\textbf{counterfactual ratio}: observed failures per dialogue divided
by what an independent-failures null with the same per-turn rate
predicts: 1.0 means failures land as if independent, above 1 means they
cluster (0.95$\times$ for SYNTHESIS; C.11).

\paragraph{Value-tracing labels.} A failed turn is \emph{traceable} when a comparable row set exists for it (SYNTHESIS: the model-emitted table; BYPASS: the typed side-channel artifact; turns with no scoreable artifact are grounding failures, excluded). A traceable failure is
\texttt{propagated} when its emitted rows are consistent with filtering
a prior turn's emission rather than the ground truth (set equality or
subset over the identity keys of prior emissions), and
\texttt{independent} when the error is fresh. When both explanations
fit, the labeler resolves to \texttt{propagated}, so the propagated
share is an upper bound; for BYPASS no model-emitted set exists, current or prior, so its zero is structural rather than a survived test.

\subsection{The depth test's degenerate null}

With SYNTHESIS's 56.9\% first-turn failure rate, the accumulation-only
null $1-(1-p)^n$ saturates at 81\% by position 2 and 92\% by position 3,
leaving at most 8--19pp of measurable headroom above the null: super-linearity of any plausible magnitude is undetectable in this regime. SYNTHESIS's marginal
hazard is flat with depth (57/58/60/39/55\% at positions 1--5); BYPASS's
position-1 hazard is 0.0\% on the main 24-dialogue corpus: all 26 of its failed turns (\S\ref{sec:failure}) sit at later positions.

\subsection{The SYNTHESIS construction's consequences, in full}

\begin{itemize}
\item \textbf{Translation error is common-mode.} The same planner, with
  the same instruction lineage, decides what the user asked for in both
  conditions, so the residual gap is attributable to delivery.
\item \textbf{Mechanical delivery makes SYNTHESIS stronger than its
  fork-true form.} In the real fork-point system the sub-agent's records
  pass through an LLM emission before reaching the planner; we remove that
  step, so our SYNTHESIS is a charitable version of the architecture it
  stands for.
\item \textbf{BYPASS's advantage is joint, and we name it.} BYPASS
  benefits jointly from synthesis bypass (records never re-typed) and
  structured-state preservation (each follow-up recomputes from the
  original set). We do not separate them: the deployed system does both,
  and no deployed pattern we know of does one without the other, so the
  comparison is the pair, not a single-factor ablation.
\end{itemize}

\subsection{Corpus composition and capture provenance}

The 408 captures decompose as 270 fixed-date award, 18 award date-range,
and 120 cash. The four date bands are \textbf{close-in} (${\approx}3$
weeks out), \textbf{near} (${\approx}3$ months), \textbf{far}
(${\approx}9$ months), and \textbf{release-edge} (${\approx}11$ months,
the edge of the booking window where award seats are first released).
Captures are restricted at capture time to non-stop and one-stop
itineraries with the pre-filter count recorded in each manifest, so the
restriction is never a silent truncation; each capture records the git
SHA that produced it, the request, and per-call vendor traffic.

\subsection{Scale}

\begin{table}[t]
\centering\small
\begin{tabular}{lr}
\toprule
Frozen captures / records & 408 / 9{,}809 \\
Routes & 36 \\
Dialogues & 24 \\
Runs per condition & 72 (24 dialogues $\times$ $k{=}3$) \\
Scored turns per condition & 339 \\
\quad of which turn-matched & 264 \\
Conditions & 2 \\
Models & 1 (Gemini 2.5 Flash) \\
\bottomrule
\end{tabular}
\caption{Main-study scale. Capture, record, and route totals are the full frozen corpus, which also serves the size and prefix instruments (Appendix~\ref{app:sizedialogues}); further families replicate at the same 339-turn scale (\S\ref{sec:family}, F.1). The headline is the full 339 scored turns per condition; the 264 turn-matched turns (identical turn sets in both instrument versions' corpora) isolate the instrument effect (\S\ref{sec:whatmoved}).}
\label{tab:scale}
\end{table}

Table~\ref{tab:scale} reproduces the study's scale.

\subsection{The four deltas, in full}

The production agent tree is deep-copied in memory with exactly four
deltas: (1) the planner's instruction replaced by a variant walked
block-by-block against production, removing only blocks that presuppose the context tool and client-side rendering; this variant, with the sampling inherited unchanged (\S\ref{sec:synthesis}), is what the abstract's ``lineage-matched'' names: the same production lineage in both arms, differing only in delivery-entailed blocks; (2) the deterministic context
tools removed from the planner's toolset; (3) each flight-search
sub-agent given a callback replacing its LLM-generated summary with
\texttt{\{total\_found, search\_params, records\}} read verbatim from
state, so the row list is delivered \emph{mechanically}; (4) those clones
dropping their output schema, which would otherwise strip the injected
records. Neither condition sets an output-token cap: the only generation parameters configured in either arm are temperature 0.1 and top\_p 0.5 (\S\ref{sec:synthesis}; Claude inherits temperature only, E.5); provider defaults apply identically across arms within each family. Each delta is entailed by the swap: in-context delivery cannot coexist with a deterministic context tool, with instruction blocks that presuppose it, or with output schemas that strip the injected rows; the abstract's ``differing only in delivery architecture'' quantifies over this one variable, implemented as these four deltas.

\subsection{Dialogue inventory}

Six arcs cover narrow-and-pick, award/cash comparison, release-edge with
truthful empty results, range/flex, multi-table refer-back, and compound
power-user filters; turn families span the filter vocabulary: time buckets, day-rollover arrival
filters, seat and program constraints, superlatives with sort and limit,
negated predicates, refer-back, and cumulative narrowing. The four variation knobs are payload swap, order permutation, follow-up addition or omission, and phrasing alternates. Dialogue
identifiers encode route and design cell
(\texttt{d02\_lhr\_fco\_close}, \texttt{d26\_fra\_bkk\_size\_m\_early};
Appendix~\ref{app:sizedialogues}).

\subsection{Operational detail moved from the body}

The fifteen scored fields are flight number, airline, booking source,
mileage cost, taxes, cabin, date, origin, destination, departure and
arrival timestamps, duration, stops, remaining seats, and cash price.
\texttt{source} is included in the identity key because the same flight
legitimately appears once per booking program. The much smaller
partial-credit gap is the expected signature: SYNTHESIS frequently emits
a \emph{nearly}-right set, which is still a wrong answer to a filter
question. Turns flagged \texttt{ordered} (limit-N superlatives and
explicit re-sorts) are scored on emitted sequence, reported as their own
axis and never folded into fidelity; mean order agreement is 95.9\% vs
58.2\% (C.3).

Every run persists a readable trace and a complete raw record (tool
calls, model texts, token usage, final state, per-turn scores). The
analysis layer computes dialogue-level bootstrap CIs (2000 resamples,
dialogue as the unit, following \citealp{schessl2026autocorrelation})
plus per-position and lag-1 hazards, cumulative failure versus the
accumulation null, accuracy by table size, and runs a value-tracing labeler that marks each traceable failure \texttt{independent} or \texttt{propagated} (definitions in C.4).

\S\ref{sec:cluster}'s cut detail, with C.5's test construction:
SYNTHESIS's excess over the accumulation-only null is negative at every
position ($-14.8$ to $-22.3$pp at positions 2--5) and its failures per
dialogue land at 0.95$\times$ the per-dialogue independence prediction (C.4); lag-1 samples are $n{=}146/121$ (SYNTHESIS, 71.9\% vs 31.4\%) and $n{=}23/244$ (BYPASS, 4.3\% vs 10.2\%; the nominal inversion, a lower hazard after a failure, is a single-failure cell, 1/23, consistent with independence rather than any protective effect).

\section{The Exhibits and the Fabrication Census in Full}
\label{app:census}

\subsection{The census decomposition}

Of 1,275 context-populated emitting turns (both corpora, production
model), 1,245 emit only delivered rows; the 30 exceptions decompose into
identifier-level corruption of real rows (25 turns: connecting segments
of multi-leg itineraries dropped or recombined, or a single garbled code, inside an otherwise-real table; e.g., a real ``\ldots, BR62'' tail swapped to ``BR68'', one run splicing 24 of its 103 emitted rows) and five turns in two pre-fix
multi-search dialogues where the \emph{queried} award table itself had
never been delivered (the context held only cash rows) and the model
produced two-row placeholder tables (\texttt{UA123}, \texttt{JL789},
\texttt{KL456}) for the missing set. Every from-nothing fabrication in
either corpus sits over a missing queried table. All exhibits are
schema-valid SYNTHESIS turns, and the census finds no third variant in
which a delivered table is present and a fabricated one replaces it.

\subsection{Exhibit 1: fabrication under grounding absence (the plausible face)}

Dialogue \texttt{d02\_lhr\_fco\_close}, sweep k1. The frozen payload is
LHR$\rightarrow$FCO on 2026-08-14, economy: \textbf{48 real records}
whose only mileage values are \{7500, 14000, 15000, 22500, 25500, 28200,
47500\} and whose booking sources are \{aeroplan, alaska, delta,
flyingblue, qantas, virginatlantic\}. The \emph{readiness gate} is a
deterministic pre-search check that all required slots are filled; its
``Ready'' verdict licenses the search dispatch. In this run it returned
``Ready'', but the search sub-agent was never invoked (the pre-fix
starvation event, \S\ref{sec:starvation}), so the context held no award
records at any turn. Asked at turn 2 for ``ones less than 10,800 points''
(ground truth 27 rows, all at exactly 7,500), SYNTHESIS emitted (identity fields shown; the records' other fields elided):

\begin{small}
\begin{verbatim}
{"flights": [
 {"flight_number": "BA234",
  "mileage_cost": "10000",
  "source": "avios"},
 {"flight_number": "VS123",
  "mileage_cost": "9500",
  "source": "virginatlantic"}],
 "summary": "I found 2 award flights for less than
  10,800 points."}
\end{verbatim}
\end{small}

None of it exists in the payload: not the flight numbers, not the mileage values, not the source \texttt{avios}; \texttt{schema\_parsed: true}. The
fabrications are \emph{plausible}: under the user's cap, on carriers
that credibly fly London$\rightarrow$Rome, with textbook flight numbers:
nothing distinguishes the reply from a genuine two-row result. And the
fiction propagates: the remaining turns filter, sort and re-emit the
invented rows. In Table~\ref{tab:ex1} (and Table~\ref{tab:ex2}), each
row's \textbf{GT is the correct table \emph{after executing} that turn's
request} under the scope shown, computed by the oracle from the frozen
payload and independent of what the run did.

\begin{table*}[t]
\centering\small
\begin{tabular}{clllc}
\toprule
t & user turn & active scope & GT & emitted \\
\midrule
1 & ``Find award flights from LHR to FCO\ldots'' & \emph{search: should deliver 48 rows; never fires} & \emph{not scored} & --- \\
2 & ``Show me ones less than 10,800 points'' & $<$10,800 pts & 27 & 2 \\
3 & ``Only morning departures please'' & + morning (composes) & 19 & 2 \\
4 & ``Of those, at least 2 seats'' & + $\ge$2 seats (composes) & 17 & 2 \\
5 & ``Give me the cheapest one'' & cheapest of the 17 & 1 (of 17) & 1 \\
6 & ``Show all the morning ones again'' & back to t3's full morning table & 19 & 2 \\
7 & ``And the 3 earliest of those'' & 3 earliest of those 19 & 3 (of 19) & 2 \\
\bottomrule
\end{tabular}
\caption{Exhibit 1 (\texttt{d02\_lhr\_fco\_close}, sweep k1): a starved
SYNTHESIS run fabricates a two-row table and then filters the fiction. GT
is the oracle's expected emission per turn; ``emitted'' is what the model
displayed.}
\label{tab:ex1}
\end{table*}

Fidelity is zero for the rest of the dialogue because the model is
filtering fiction; on the same starved dialogue, BYPASS emits zero rows, a visible failure.

\subsection{Exhibit 2: fabrication under starvation}

Dialogue \texttt{d23\_jfk\_lhr\_relax\_close}, sweep k2. At turn 1 the
readiness gate returned ``Ready'' but the award search sub-agent was
never invoked, so \textbf{no records ever entered the context}: the frozen payload this dialogue should have delivered holds 74 rows (the
same no-search event appears in BYPASS runs; see
\S\ref{sec:integrity}). Asked at turn 2 for flights under 30k points,
SYNTHESIS made zero tool calls and emitted two placeholder-shaped records (\texttt{UA123 / 25000 / united\_miles}, \texttt{DL789 / 28000 / delta\_skymiles}): round point values, flight numbers absent from the frozen payload, non-production source strings. Asked at turn 4 to widen the cap (``Actually, up to 87,499 points is fine!''), the model does not return to any source: it \textbf{generates three additional fictional rows} (\texttt{BA456 / 35000 / avios}, \texttt{VS101 / 40000}, \texttt{AA200 / 80000}), one flight per major carrier, \textbf{to fill the widened window} (Table~\ref{tab:ex2}).

\begin{table*}[t]
\centering\small
\begin{tabular}{c p{3.0cm} p{4.7cm} l p{2.7cm}}
\toprule
t & user turn & active scope & GT & emitted \\
\midrule
1 & ``Find award flights from JFK to LHR\ldots'' & \emph{search: should deliver 74 rows; never fires} & \emph{not scored} & --- \\
2 & ``less than 30k points'' & $<$30k pts (over the full 74) & 23 & 2 (fabricated) \\
3 & ``morning departures'' & + morning (composes) & 6 & 2 (the same fiction re-emitted) \\
4 & ``Actually, up to 87,499 points is fine!'' & cap \textbf{relaxed} to $\le$87,499 over the full table (morning no longer active) & 44 & 5 (more fiction) \\
5 & ``cheapest one'' & cheapest of the 44 & 1 (of 44) & 1 \\
\bottomrule
\end{tabular}
\caption{Exhibit 2 (\texttt{d23\_jfk\_lhr\_relax\_close}, sweep k2):
starved SYNTHESIS fabricates placeholder rows, then fabricates more to
fill a relaxed cap. The t4 view grows past t2's 23 because the user is
backing off the budget set at t2: ``up to 87,499 points is fine'' asks about
everything under the new cap, not just the morning subset they had
drilled into (the scope semantics of \S\ref{sec:authoring}). Contrast t3,
which narrows, and Exhibit 1's t6, which returns to an earlier view.}
\label{tab:ex2}
\end{table*}

Under the same starvation, BYPASS emits zero rows (visibly empty, so the user retries) where SYNTHESIS emits a complete, plausible,
schema-valid table that is entirely false.

\section{Integrity Detail}
\label{app:integrity}

\subsection{The three ground-truth defects, in full}

(1) An unscoreable turn: ``Which day is cheapest?'' has no defined
aggregate (Appendix~\ref{app:authoring}) and was retired under the
\S\ref{sec:authoring} rule, replaced with an unambiguous window-spanning
filter turn (``Morning departures across all the days''). (2) Stale
downstream ground truth: after the replacement, later turns' ground truth
was not re-composed, penalising the deployed condition for being right:
one turn demanded 64 rows (date only) while BYPASS's 9 (morning $\cap$
date) was correct under our own scope convention. The arc's scored-turn count drops from 27 to 25 in the correction: for the two payloads with zero morning departures (d16, d17) the replacement turn is itself the seats filter, so each dialogue's later standalone ``Of those, at least 2 seats'' turn became a no-op duplicate and was removed in the re-composition; both conditions had answered both turns correctly, so the drop favours neither side. (3) A hardcoded day pick (range-start + 2) on one payload selected a day with zero
morning departures, unscoring every downstream turn.

The production bug the same arc captured: a date-window argument the
context tool did not accept was silently dropped by the framework, so a
user asking for a week received one day (2 rows instead of 6); post-fix
the same turns return 467 / 6 / 6 rows, exact. A third production fix from the same study: duplicate cash itineraries had been reaching users (61 rows across 21 captures; now zero). The two ground-truth semantics corrections of \S\ref{sec:intro} are \S\ref{sec:authoring}'s refer-back and constraint-revision scope rulings.

\subsection{The starvation mechanisms}

For BYPASS, the production prompt's worked examples taught the model to
answer instead of searching; the fabrication fix (\S\ref{sec:postfix})
repaired it, taking the rate on affected dialogues from 14.2\% to 1.4\% at the time of the production fix (on the post-fix main corpus the BYPASS rate is 0.0\%; \S\ref{sec:starvation}).
For SYNTHESIS, \texttt{output\_schema} biases the planner toward emitting
a schema-valid ``searching\ldots'' object in place of the tool call, an
ADK-level behaviour that an ordering instruction mostly but not fully
counteracts; the residual post-fix 9.7\% is schema pressure winning
over the instruction. The schema-versus-tool-call tension is documented in the framework's own tracker: declared output schemas are unreliably honoured when tools are present \citep{adk2025schematools}, and an input-schema variant suppresses an inner agent's tool calls entirely \citep{adk2026schemaskip}.

\subsection{The position question: the three instruments' design}

The size ladders (\S\ref{sec:size}) counterbalance position: each size
band appears at both an early and a late position. An observational
regression over all 527 scored SYNTHESIS turns of the main 24-dialogue corpus and the size ladders
(logistic, correct $\sim \log_2(\text{size}) + \text{position}$) puts the
size term at $\chi^2{=}83.3$ ($p{=}7{\times}10^{-20}$) with position
controlled, leaving a small residual position term ($\chi^2{=}6.1$,
$p{=}0.013$) that observational data cannot adjudicate (accumulated context grows with position by construction), and that the Latin square
was designed to settle. The square: two routes, three delivered sizes
(9/11, 70/74, 174/194 rows) each appearing at each dialogue slot exactly
once per route, byte-identical scored turns in every slot, oracle-verified
identical ground truth for a size at every slot.

The square's cell pattern hints at one real secondary texture: small-table accuracy drifts down when large tables precede it in context
(92/75/58\% across slots) while large-table accuracy does not worsen, a mild accumulated-context tax, far from accounting for the size effect.

\subsection{The headline movement, accounted in full
(\S\ref{sec:whatmoved})}

The BYPASS side: removing starvation (0.0\% post-fix) and the
fabrication-teaching prompt lifted BYPASS from 81.4\% to 91.3\%
($+9.9$pp). The SYNTHESIS side: two of the eight fixes (timestamp
canonicalisation and tie-break re-authoring) removed penalties only
SYNTHESIS could incur and were expected to \emph{raise} it by roughly
4pp, yet SYNTHESIS fell from 47.7\% to 41.7\% ($-6.0$pp). Per-dialogue
movement is large in both directions ($+41.7$ to $-46.7$pp on individual
dialogues), consistent with SYNTHESIS's high run-to-run variance, but the
aggregate move exceeds the pre-fix corpus's own sweep-to-sweep spread and we do
not have a mechanism for it.

\subsection{Sampling-parity ablation (Limitations)}
\label{app:parity}

Claude runs with top\_p unset because Anthropic's API rejects temperature
and top\_p together. A Gemini ablation (the only family where the combination is settable) measured clearing top\_p at $-15.7$pp for BYPASS and $-5.9$pp for SYNTHESIS; if the direction transfers across families, the parity break biases against BYPASS and the Claude gap is conservative.

\section{Results Detail}
\label{app:results}

\subsection{Cross-family full matrix (\S\ref{sec:family})}

Table~\ref{tab:family} gives the per-condition cells behind \S\ref{sec:family}'s paired gaps ($n{=}339$
scored turns per cell: all 24 dialogues; Wilson 95\% CIs on cells,
dialogue-level bootstrap CIs on gaps; the cell intervals pool turns and are descriptive; inferential gap claims rest on the dialogue-level CIs). Extending the repeats from three to five sweeps for the two non-Gemini families (GPT-4o-mini and Claude Haiku 4.5; $n{=}565$ per cell) moves no cell by more than 1.1pp and no gap by more than 1.4pp.

\begin{table*}[t]
\centering\small
\begin{tabular}{llll}
\toprule
model & BYPASS & SYNTHESIS & gap \\
\midrule
Gemini 2.5 Flash & 92.3\% [89.0, 94.7] & 45.7\% [40.5, 51.0] & \textbf{+46.6pp} [+35.7, +57.1] \\
Gemini 2.5 Pro & 92.6\% [89.3, 95.0] & 72.0\% [67.0, 76.5] & \textbf{+20.6pp} [+11.7, +31.4] \\
GPT-4o-mini & 67.0\% [61.8, 71.8] & 17.4\% [13.7, 21.8] & \textbf{+49.6pp} [+39.2, +60.2] \\
Claude Haiku 4.5 & 82.9\% [78.5, 86.5] & 27.1\% [22.7, 32.1] & \textbf{+55.8pp} [+41.1, +69.9] \\
\bottomrule
\end{tabular}
\caption{Cross-family replication at full-matrix scale ($n{=}339$ scored turns per cell, k1--3; Wilson 95\% CIs on cells, dialogue-level bootstrap CIs on gaps).}
\label{tab:family}
\end{table*}

\subsection{Trend-test robustness (\S\ref{sec:size})}

Because repeats within a route are correlated, the decreasing top-K trend
was also verified under route-level cluster bootstrap: it survives,
appearing in 2000 of 2000 resamples, i.e., $p$ below the bootstrap's resolution floor ($p<5{\times}10^{-4}$).

\subsection{Family failure shapes, specifics (\S\ref{sec:family})}

GPT's garbling includes type-broken fields (``21h 55m'' where the schema
wants an integer) and responses that drop required fields (a
\texttt{flights} array with no \texttt{summary}, the mirror image of
Claude's omission), including 13 runs with no scoreable output at all, four of them because the re-emitted tables inflated the transcript past the model's 128k context window; BYPASS, whose context never carries the table, has no such run.
Claude's omission mode appeared on thirty-five runs: twenty-nine with an intact prose summary
counting the delivered table correctly (``Found 48 award
flights\ldots'') and six emitting an empty object. Omission is even cheap (4.0\,s per Claude synthesis
turn; not emitting the table is fast), where garbling is not (16.0\,s per
GPT synthesis turn). A three-dialogue pilot had suggested Claude's
synthesis was markedly stronger (64.7\% there); that was an artifact of
the pilot's small tables, which rarely trigger the omission mode. GPT's
ladder zero at K=5 (0/6) against 33\% at K=10 is a real non-monotonicity
of its garble-dominated mode. A power note on size-flat BYPASS: a zero-failure cell excludes, with 80\% probability, only a true per-turn decay of at least 8.5pp at $n{=}18$ (prefix-ladder cells) and 23.5pp at $n{=}6$ (family-ladder cells); the exclusion of a large-size decay rests on pooling across the four instruments (93/93 prefix ladder, 140/144 family ladder, 238/240 date-varied ladder, 106/108 Latin square), where every BYPASS miss is starvation-shaped (empty delivery, no rows reaching state) and none is size-graded.

\subsection{The tier's residual failures and per-answer costs
(\S\ref{sec:family}, \S\ref{sec:cost})}

\begin{figure}[t]
  \centering
  \includegraphics[width=\columnwidth]{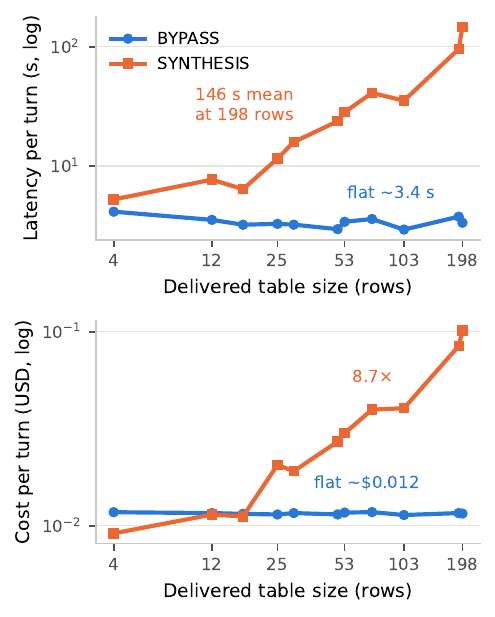}
  \caption{Per-turn latency (top) and cost (bottom) against the actual delivered table size, over the size ladder's scored filter turns. SYNTHESIS
  scales from ${\sim}5$s to a 146s mean turn and, at the largest size, 8.7$\times$ BYPASS's per-turn cost;
  BYPASS is flat at ${\sim}3.4$s and ${\sim}\$0.012$ at every size. At the smallest tables SYNTHESIS is marginally cheaper, crossing above BYPASS by 25 rows. (\S\ref{sec:cost}'s 4.9s/31.7s are all-turns means over the full grid; this plot's means cover the ladder's scored filter turns only. \S\ref{sec:size}'s design labels are post-seat-filter counts; this
  figure uses the true delivered sizes.)}
  \label{fig:cost}
\end{figure}

Gemini 2.5 Pro's full-corpus cells are in the cross-family matrix (F.1); a Pro synthesis turn averages 35.5\,s against Flash's 22.8\,s (scored filter turns, the same turn class Figure~\ref{fig:cost} plots; \S\ref{sec:cost}'s 31.7\,s is the all-turns mean; 1.6$\times$) and \$0.141 against Flash's \$0.028 (5$\times$), twelve times BYPASS-on-Flash's \$0.012. Residual synthesis failures carry the size signature of \S\ref{sec:size} (in the tier pilot runs: near-miss truncation on 40--77-row tables, 77 delivered $\rightarrow$ 74/76 emitted; collapse at 138 $\rightarrow$ 0 and 76 emitted).

 On
the content-controlled ladder Pro's correct full-payload answers cost
\$0.67 each (3 of 6 correct at a mean \$0.336 per turn) against BYPASS's
flat \$0.012. A pre-fix probe on five failure-selected dialogues
agreed with the tier comparison's direction: Flash-BYPASS 60.0\% vs
Pro-SYNTHESIS 52.0\%.

\subsection{Failure anatomy (\S\ref{sec:failure})}

Beyond truncation, SYNTHESIS over-emits by carrying a filter it should
have dropped. On the deployed Gemini 2.5 Flash, BYPASS's 22 wrong-argument failures are translation and scope errors; corpus-wide, its applied arguments differ from the authored spec on 27.2\% of argument-recoverable scored turns (92/338 of 339), but 70 of those 92 argument sets
(76\%) still select exactly the right rows: argument-level noise
absorbed by the deterministic evaluator, where synthesis would have
re-typed every field.

\begin{table*}[t]
\centering\small
\begin{tabular}{lrrrr}
\toprule
model & BYPASS turns correct & argument divergence & of which harmless & failed turns from wrong args \\
\midrule
GPT-4o-mini & 67.0\% & 45.6\% (150/329) & 41\% (61/150) & 79\% (89/112) \\
Claude Haiku 4.5 & 82.9\% & 32.0\% (103/322) & 63\% (65/103) & 66\% (38/58) \\
Gemini 2.5 Flash & 92.3\% & 27.2\% (92/338) & 76\% (70/92) & 85\% (22/26) \\
Gemini 2.5 Pro & 92.6\% & 31.2\% (105/336) & 82\% (86/105) & 76\% (19/25) \\
\bottomrule
\end{tabular}
\caption{Cross-family translation error over scored BYPASS turns ($n{=}339$ per model; rows ordered by ascending BYPASS accuracy). Divergence = applied filter arguments differ from the authored spec; denominators are turns with trace-recoverable applied arguments (the shortfall of 10/17/1/3 turns is all failures). Harmless = divergent arguments that still select exactly the right rows, so divergent minus harmless = the wrong-argument failures. Last column: wrong-argument failures over all failed turns; the remainder are 13/3/3/3 failures with right arguments plus the 10/17/1/3 unrecoverable, closing each total (112/58/26/25).}
\label{tab:xfam-translation}
\end{table*}

Wrong-argument failures order exactly inversely with the four BYPASS
accuracies (Table~\ref{tab:xfam-translation}), including the Pro/Flash
flip at the top, and the harmless share rises with BYPASS accuracy (41\% to 82\%):
BYPASS's floor is translation quality (how well a model maps each
follow-up's intent into filter arguments, \S\ref{sec:deployed}),
measured.

\section{Extended related work and system detail}
\label{app:extended}

\subsection{Extended related work}

The body's \S\ref{sec:related} keeps the closest works; the full positioning:

\paragraph{Structured output: validity is not correctness.} Constrained
decoding guarantees shape, not content: format restriction degrades
reasoning \citep{tam2024freely}, generation shifts onto lower-confidence
paths \citep{schall2025hidden}, validity hits 100\% while accuracy falls
and wrong-but-valid outputs dominate \citep{ray2026constraint}, task
accuracy coexists with zero format accuracy \citep{galeone2026usable};
grammar-constrained infrastructure
\citep{willard2023outlines,dong2024xgrammar} and schema-compliance
benchmarks \citep{geng2025jsonschemabench} show the shape guarantee is
real. \citet{hoang2026structured} measured the resulting structured
hallucination in a production travel agent single-turn, introducing the
field-fidelity metric and zero-out policy we adopt. This paper compares
two delivery architectures on a new dataset in a multi-turn design
(results are not compared to that study's), showing the failure survives
into a \emph{multi-turn agentic pipeline over real vendor records},
where corrupted output re-enters the context and is filtered again.

\paragraph{Multi-turn agent evaluation and error compounding.} Tool-use
work established reliable invocation
\citep{yao2023react,schick2023toolformer,patil2024gorilla}; our subject
is what happens to the returned data afterwards. $\tau$-bench established
multi-turn tool-agent evaluation \citep{yao2024taubench}, MINT showed
multi-turn ability is not predicted by single-turn \citep{wang2024mint},
a 39\% average single- to multi-turn drop with poor recovery is
documented \citep{laban2025lost}, and self-conditioning, in which past errors raise later error rates, is the closest mechanism match
\citep{sinha2025illusion}; failure taxonomies \citep{zhu2025fail},
stability analyses \citep{xiong2025vulnerable}, reliability framing
\citep{rabanser2026reliability}, a Tool-Skip category matching our
secondary observation \citep{soni2026toolfailbench}, and injected
corruption producing over-trust and reduced recovery \citep{toolmaze2026}
situate the problem. Closest are three unreviewed preprints:
AgentProp-Bench (synthetic tasks, injected errors) defers multi-turn injection (our question) to future work \citep{gurram2026agentprop};
``Beyond Task Completion'' reports 27--78\% of $\tau$-bench successes as
\emph{corrupt} \citep{cao2026corrupt}: procedural integrity on $\tau$-bench's mock domains (airline and retail), where ours is field- and row-level data fidelity;
``Reason Less, Verify More'' finds, on a budget agent, 78\% of failures are silent wrong-state and gates the \emph{write} side \citep{reddy2026gates}, where
ours bypasses the \emph{read} side. All index degradation by turn count
or injected fault on simulated or injected data; ours is indexed by
delivered record-set size, on real data, and the corruption is
self-inflicted.

\paragraph{Architectures that keep data out of the model.} CodeAct routes
actions through executable code \citep{wang2024codeact}; ``heavy brain,
precise hands'' designs pair LLM planning with a deterministic backend
\citep{zeng2026rethinking}; trace-level compliance verification adds a
deterministic gate \citep{elkoussy2026agentltl}; the prompt-injection
literature converges on context-minimization
\citep{beurer2025designpatterns}. Our mechanism is in this family; its
practitioner form exists in LangGraph's custom stream mode, the Vercel AI
SDK's data parts, and the AG-UI protocol, but \textbf{no framework we surveyed ships
typed structured delivery as the default end-to-end rendering path}, and
the public motivation is rendering predictability, never correctness
\citep{copilotkit2026genui}. Survey method: for each framework (LangGraph, Vercel AI SDK, CrewAI, Microsoft Agent Framework, OpenAI Responses API, CopilotKit/AG-UI; documentation accessed 28 July 2026) we read the current public streaming and rendering documentation and asked one question: does structured tool output reach the client renderer without re-entering the model's context on the documented default path? Default means the quickstart path, not an available opt-in.

\paragraph{Evaluation methodology.} Turns within a dialogue are not
independent, and turn-level significance claims frequently fail
cluster-robust correction \citep{schessl2026autocorrelation}; every
gap interval here therefore resamples \emph{dialogues}, not turns,
a choice survey work also supports \citep{survey2025multiturn}.

\paragraph{Reasoning inside dialogue.} Closest in setting is BOULDER
\citep{kartac2026boulder}: reasoning over MultiWOZ-derived travel
domains, isolated and embedded in dialogue, with a substantial
isolated-to-dialogue drop attributed largely to multi-turn framing.
BOULDER measures whether the model \emph{computes correctly over records it was given}; we measure whether the record set \emph{survives
delivery}; it has no synthesis-bypassing arm and no row- or field-level
fidelity metric, and its ablations cannot separate the assistant's dual
generation role from longer context; the generation burden can only be
removed architecturally.

\subsection{Deployed-system detail}

\emph{Award} inventory is seats bookable with loyalty points, priced per booking program, so the same flight legitimately appears once per program; \emph{cash} is ordinary priced tickets. The search sub-agents perform hub routing, deduplication and canonicalisation before writing the record set to session state.

\end{document}